# 3D Conditional Image Synthesis of Left Atrial LGE MRI from Composite Semantic Masks

Yusri Al-Sanaani[1], Rebecca Thornhill[1,2], and Sreeraman Rajan[1]
[1]Systems and Computer Engineering, Carleton University, Ottawa, Canada
[2]Department of Radiology, University of Ottawa, Ottawa, Canada
yusrialsanaani@cmail.carleton.ca

***Abstract*—*Segmentation of the left atrial (LA) wall and endocardium from late gadolinium-enhanced (LGE) MRI is essential for quantifying atrial fibrosis in patients with atrial fibrillation. The development of accurate machine learning-based segmentation models remains challenging due to the limited availability of data and the complexity of anatomical structures. In this work, we investigate 3D conditional generative models as potential solution for augmenting scarce LGE training data and improving LA segmentation performance. We develop a pipeline to synthesize high-fidelity 3D LGE MRI volumes from composite semantic label maps combining anatomical expert annotations with unsupervised tissue clusters, using three 3D conditional generators (Pix2Pix GAN, SPADE-GAN, and SPADE-LDM). The synthetic images are evaluated for realism and their impact on downstream LA segmentation. SPADE-LDM generates the most realistic and structurally accurate images, achieving an FID of 4.063 and surpassing GAN models, which have FIDs of 40.821 and 7.652 for Pix2Pix and SPADE-GAN, respectively. When augmented with synthetic LGE images, the Dice score for LA cavity segmentation with a 3D U-Net model improved from 0.908 to 0.936, showing a statistically significant improvement ($p < 0.05$) over the baseline.. These findings demonstrate the potential of label-conditioned 3D synthesis to enhance the segmentation of under-represented cardiac structures.*

## I. Introduction

Late gadolinium-enhanced magnetic resonance imaging (LGE-MRI) may aid guidance of ablation therapy in patients with atrial fibrillation. Accurate segmentation of the left atrial (LA) wall is a critical first step in post-processing LGE-MRI [1]. Manual tracing of LA myocardium and scars on 3D LGE MRI is highly labour-intensive. Advances in deep learning have significantly improved the segmentation of the left atrium cavity in LGE-MRI, achieving a Dice score of 0.932 on the benchmark dataset using two-stage U-Net model [2]. However, segmentation of the thin LA wall (where fibrotic remodeling occurs) remains challenging and less explored than in conditions involving the left ventricular wall, which is three or four times as thick as the LA wall. The atrial myocardial wall is extremely thin (2-3 mm) and encloses a complex chamber with pulmonary vein and mitral valve openings. Furthermore, the LA wall and scar occupy far fewer voxels than the blood pool or background, causing a severe class-imbalance problem for segmentation networks trained to optimize for the larger classes [1], [2], [3]. Another barrier in LA wall segmentation is the scarcity of large, annotated training datasets. These challenges motivate the use of synthetic data augmentation to improve generalization, especially for underrepresented and anatomically complex structures, such as the left atrial wall. Advances in generative adversarial networks (GANs) and diffusion models enable the synthesis of realistic images conditioned on structural priors, such as segmentation masks, class labels, or images from other modalities [4], [5], [6]. The seminal Pix2Pix framework by Isola et al. first demonstrated that a U-Net generator conditioned on segmentation maps can produce realistic images [7]. In medical imaging, this idea was quickly adopted: e.g., Shin et al. (2018) trained a 2D Pix2Pix model to translate brain segmentation maps into MR images and demonstrated that a network trained on a mix of real and synthetic MRIs achieved higher tumour segmentation accuracy than a network trained on real images alone [4].

A major innovation in conditional GANs was the spatially adaptive denormalization (SPADE) layer proposed by Park et al. [8]. In brief, SPADE injects semantic label information at multiple layers of the generator through conditional normalization, thereby helping to preserve exact shape details from the label map. Recently, SPADE-based generators have been adopted in medical image synthesis. Lustermans et al. (2022) reported that 2D SPADE GAN–based augmentation, conditioned on myocardium and scar labels, improved left ventricular scar segmentation (Dice score: 0.72 to 0.78, $p < 0.01$) [5].

Whereas GANs rely on adversarial training, diffusion models learn to iteratively refine random noise into a realistic image by reversing a noising process. This is typically accomplished using a U-Net to predict denoised images at each step [9], [10]. Latent diffusion models (LDM) proposed by Rombach et al. mitigate the computational demand by performing diffusion in a lower-dimensional latent space rather than pixel space [11]. LDMs have also been adapted to support semantic synthesis by incorporating SPADE-style conditioning; Zhou et al. introduced a variant where semantic labels modulate decoder activations through spatially adaptive normalization layers, improving realism and structural fidelity [12]. Ruschke et al. demonstrated that adding SPADE-conditioned latent diffusion–generated MRIs to real data improved segmentation accuracy, improving the Dice score from 0.66 to 0.72 [13].

In summary, conditional generative models have evolved from 2D Pix2Pix GANs to more advanced architectures, with notable gains in realism and structural fidelity. However, most existing studies are limited to 2D synthesis or target anatomically simpler structures, such as the brain or left ventricle, often relying on sparse labels and lacking contextual conditioning. In contrast, 3D LA synthesis requires preserving cross-slice continuity and modelling thin, complex structures accurately. Recent studies emphasize the importance of 3D volumetric synthesis in medical imaging to preserve spatial context and anatomical continuity [10], [14].

In this paper, we evaluate three distinct 3D conditional generative models for LGE MRI synthesis and their impact on LA segmentation. We, therefore, adapt three 3D architectures – Pix2Pix, SPADE-GAN, and an improved latent diffusion model with SPADE conditioning (SPADE-LDM) – spanning both GAN- and diffusion-based approaches. We utilize 3D quantitative evaluation metrics to assess the quality of images generated by these models. To our knowledge, this is the first comparison of these 3D models for LA LGE MRI synthesis.

Prior work shows that when generators are conditioned on isolated organ masks, they often fail to produce realistic surroundings [15], [16]. We address this challenge of anatomical context by introducing composite semantic label maps that include LA wall and LA blood pool (cavity) masks along with contextual anatomical structures derived via unsupervised clustering. Conditioning on these composite label maps enables the models to synthesize LGE MRI LA within a realistic cardiac environment from these composite labels. Finally, we demonstrate the efficacy of synthetic augmentation by training a segmentation model for the LA cavity with and without synthetic images and report the improvement.

## II. METHOD

### A. Data

We utilize the 2018 MICCAI Atrial Segmentation Challenge dataset, which includes 154 3D LGE-MRI volumes (from 60 atrial fibrillation patients, each scanned before and/or after ablation therapy) with expert-labeled LA endocardial and epicardial borders [2]. The original volumes have a matrix size of 576×576×88 or 640×640×88, with an isotropic resolution of 0.625×0.625×0.625 mm³. We resample them to a standardized 256×256×64 matrix with a voxel spacing of 1.0×1.0×1.0 mm³ for consistency and computational efficiency. We split the 154 patient volumes patient-wise as follows: 144/10 cases for training and validation of all generative models and 124/30 cases for training and testing the downstream segmentation network.

### B. Composite Semantic Label Maps

In addition to endocardial and epicardial (wall) masks, we introduce auxiliary labels representing surrounding tissue regions segmented via intensity-based clustering to provide anatomical context during generation. Prior studies reported that missing background context could result in unrealistic anatomies [15], [16]. To address this, we incorporate surrounding anatomical structures into the conditioning labels by generating composite multi-label masks that combine expert annotations with intensity-based k-means clustering, as illustrated in Algorithm 1. We select k-means clustering for its balance of efficiency and anatomical clarity, as well as its scalability to large 3D volumes, compared to spectral, hierarchical, or Gaussian mixture models. Gaussian smoothing (σ=1) is applied to reduce noise and improve spatial coherence. Let 3D LGE MRI volume be denoted as $\mathcal{V} \in \mathbb{R}^{H\times W\times D}$, with binary expert masks (endo and wall) $\mathcal{M}_e, \mathcal{M}_w \in \{0,1\}^{H\times W\times D}$, a voxel index set: $\Omega = \{1,\dots,H\} \times \{1,\dots,W\} \times \{1,\dots,D\}$, and clustering function $\mathcal{C}_K$. The clustering $\mathcal{C}_K$ is applied to each $\mathcal{V}$ to delineate tissue regions ($\mathcal{L}_c$, cluster labels) according to their respective signal intensity. To ensure anatomical consistency, cluster labels overlapping expert-defined masks ($\mathcal{M}_e$ and $\mathcal{M}_w$) are set to zero. We assign fixed labels for endocardium and wall (1 and 2, respectively). The background cluster ($b$) is identified dynamically by its intersection with zero-intensity voxels and excluded. The remaining voxels receive cluster-derived labels, beginning with 3, where $j(k)$ is the index of label $k$ in U, and $r(k)$ is the remapped label. This hybrid labeling ($\mathcal{L}_{composite}$) enables efficient extension of sparse annotations into semantically enriched masks for generation.

Algorithm 1: Generating composite semantic label maps

1: ***for*** *each* $\mathcal{V}, \mathcal{M}_e$ *and* $\mathcal{M}_w$ *do:*

2: $\mathcal{L}_c = \mathcal{C}_K(\mathcal{V}) \in \{0,1,\dots,K-1\}^{H\times W\times D}$ *(initial cluster map)*

3: $b = min\{k \in \{0,\dots,K-1\} | \exists\, p \in \Omega: \mathcal{L}_c(p) = k \text{ and } \mathcal{V}(p) = 0\}$

4: $\mathcal{L}_c^*(p) = \begin{cases} 0, & \mathcal{M}_e(p) = 1 \text{ or } \mathcal{M}_w(p) = 1 \\ \mathcal{L}_c(p), & otherwise \end{cases} \quad \forall p \in \Omega$

5: $U = (u_0, \dots, u_{m-1}) = sort(\mathcal{L}_c^*(\Omega))$

*for every* $k \in U$ *do:* $j(k) = \{j | u_j = k\}$

6: $r(k) = 2 + j(k), \quad k \in U$

*end for*

7: $\mathcal{L}_{composite}(p) = 0 + 1 \cdot \mathcal{M}_e(p) + 2 \cdot \mathcal{M}_w(p) + \sum_{\substack{k\in U \\ k\neq b}} r(k)\, 1[\mathcal{L}_c^*(p) = k] \quad \forall p \in \Omega$

8: *Save* $\mathcal{L}_{composite}$ *and* $\mathcal{V}$

9: ***end for***

### C. Generative Model Architectures

We investigate three 3D conditional generative models for the synthesis of LGE MRI volumes from composite semantic label maps.

We extend the 2D Pix2Pix framework [7] to 3D using a U-Net generator with five downsampling/upsampling layers and skip connections. Linear upsampling replaces transposed convolutions to reduce checkerboard artifacts [17]. The generator takes a single-channel semantic label map and produces a synthetic MRI. The 3D PatchGAN discriminator, receiving a 2-channel input (label map + real/synthetic image), uses LeakyReLU and batch normalization. The training uses adversarial loss and voxel-wise L1 loss. Adversarial loss enhances realism by encouraging outputs indistinguishable from real data. L1 loss promotes voxel-level fidelity by penalizing absolute intensity differences. We train for 400 epochs with a batch size of 1 using Adam optimizer with learning rate (LR) of $2e^{-4}$ and moment decay rates ($\beta_1$=0.5, $\beta_2$=0.999).

We implement a 3D SPADE-GAN, following Park et al. [8], where a U-Net generator integrates SPADE blocks that inject spatially adaptive affine transforms from one-hot encoded label maps at each decoder layer. Group normalization is used within SPADE blocks. The model follows a variational autoencoder (VAE)-GAN architecture, where latent codes are sampled from a learned posterior distribution to encourage diverse image synthesis by introducing stochastic variation across samples. The decoder includes four SPADE residual blocks. Dual discriminators (full and half resolution) are used, each with three convolutional layers and instance normalization. Training includes adversarial, feature-matching (stabilizes the model training by aligning intermediate discriminator features), and perceptual loss (via MedicalNet ResNet-50) to encourage high-level structural similarity. The latent space is regularized via Kullback-Leibler (KL) divergence to follow a prior distribution for sample diversity. We train for 200 epochs using Adam optimizer (LR = $2e^{-5}$ for the generator and LR = $4e^{-5}$ for the discriminator) with cosine LR scheduling and trilinear upsampling.

SPADE-LDM integrates SPADE conditioning with a latent diffusion framework. It comprises a 3D convolutional autoencoder and a latent-space diffusion U-Net. Fig. 1 illustrates the complete generative process, where an autoencoder is first trained to reconstruct MRI volumes under semantic label maps conditioning, and its encoder is later used to produce latent representations for the diffusion model, which applies iterative denoising to synthesize realistic outputs guided by the same semantic label maps. The encoder utilizes four residual downsampling blocks, with the bottleneck having eight channels. The autoencoder is trained with L1, perceptual (MedicalNet ResNet-50), and KL divergence losses. A 3D PatchGAN discriminator with R1 gradient penalty (penalizes the discriminator's sensitivity to real inputs by minimizing the squared norm of its gradients) is added post-warm-up to improve training stability and prevent discriminator overfitting. Optimization uses mixed-precision and Adam optimizer (LR = $2e^{-4}$ for the generator, and LR = $4e^{-4}$ for the discriminator) for 200 epochs. The diffusion U-Net is conditioned via SPADE at each denoising step. Inputs include the noised latent, time-step embedding, and downsampled label map. We adopt the latent denoising diffusion probabilistic model (DDPM) mean-squared-error loss of Ho et al. [9] applied to VAE latent with cosine noise schedule, as in (1). The diffusion model is trained to predict the Gaussian noise ε that corrupts the latent code, conditioned on the semantic masks through SPADE.

$$\mathcal{L}_{denoise}(\theta) = \mathbb{E}_{t,z,\epsilon}[||\epsilon - \epsilon_\theta(z_t, t, c)||^2] \tag{1}$$

where x is the MRI volume, z =E(x) its latent code produced by the encoder q(z|x); t is a diffusion time-step sampled uniformly from 1:T; ε∼N(0, I) is standard Gaussian noise; $\beta_t$ is the cosine noise schedule with $\alpha_t = 1-\beta_t$ and $\bar{\alpha}_t = \prod_{s=1}^{t} a_s$; $z_t = \sqrt{\bar{a}_t} z + \sqrt{1-\bar{a}_t}\epsilon$, which is the noisy latent given to the U-Net; $c$ denotes the down-sampled semantic label map injected by SPADE; and θ predicts the added noise.

A segmentation-consistency loss is applied by predicting masks from denoised samples using a frozen 3D U-Net, pre-trained on the same dataset (real only, as discussed later). Predicted masks are compared to ground truth using a combined Dice + cross-entropy loss, propagating gradients across the denoising path. This loss enforces anatomical correctness by aligning output masks with ground truth. We follow the shape-consistency strategy proposed by Zhang et al.[18], replacing the pure cross-entropy term with a balanced Dice + CE mix to accommodate class imbalance in left-atrium masks, as in (2).

$$\mathcal{L}_{SC} = E_x\left[\left(1 - Dice\left(\hat{s}(x), s^*\right)\right) + CE\left(\hat{s}(x), s^*\right)\right] \tag{2}$$

where x is the decoded MRI volume produced at the current diffusion step; $\hat{s}(x)$ is the mask predicted by a frozen 3D U-Net; $s^*$ is the ground-truth label map.

Training employs an Adam optimizer (LR=$1e^{-4}$) for 200 epochs. To enable classifier-free guidance (CFG), 10% of training steps use null label maps, allowing the model to learn both conditional and unconditional paths. At inference, predictions are blended with a guidance weight of 1.5 to enhance controllability, following [9].

For downstream evaluation, we utilize a single-stage 3D U-Net to segment the LA blood pool (cavity). The network employs a five-level encoder-decoder with residual units, 3D convolutions, and instance normalization. A dropout rate of 0.2 is applied. The training utilizes a composite loss (soft Dice + cross-entropy) with class weighting that favours the LA blood pool to address foreground-background voxel imbalance, where weights are computed inversely proportional to voxel counts. Here, the AdamW optimizer (LR = $1e^{-4}$) with L2 regularization ensures stability. We train under two settings: (i) 124 real MRI volumes with masks and (ii) an augmented set combining 124 real and 124 synthetic volumes (1:1) with 200 epochs. Segmentation performance is evaluated on a hold-out set of 30 only real LGE MRI volumes using the Dice score.

All generative and segmentation models are implemented using PyTorch and MONAI and trained on NVIDIA Tesla V100-SXM2 GPUs with 32 GB memory. We apply data augmentation during training, including spatial transforms (random flips, affine transformations, and elastic deformations) and intensity augmentations (Gaussian noise, blurring, gamma correction, and bias field simulation), applied only to MRI volumes. Hyperparameters were selected based on common practices in prior studies [7], [8], [9], with minor tuning to ensure convergence and visual quality stability on our dataset.

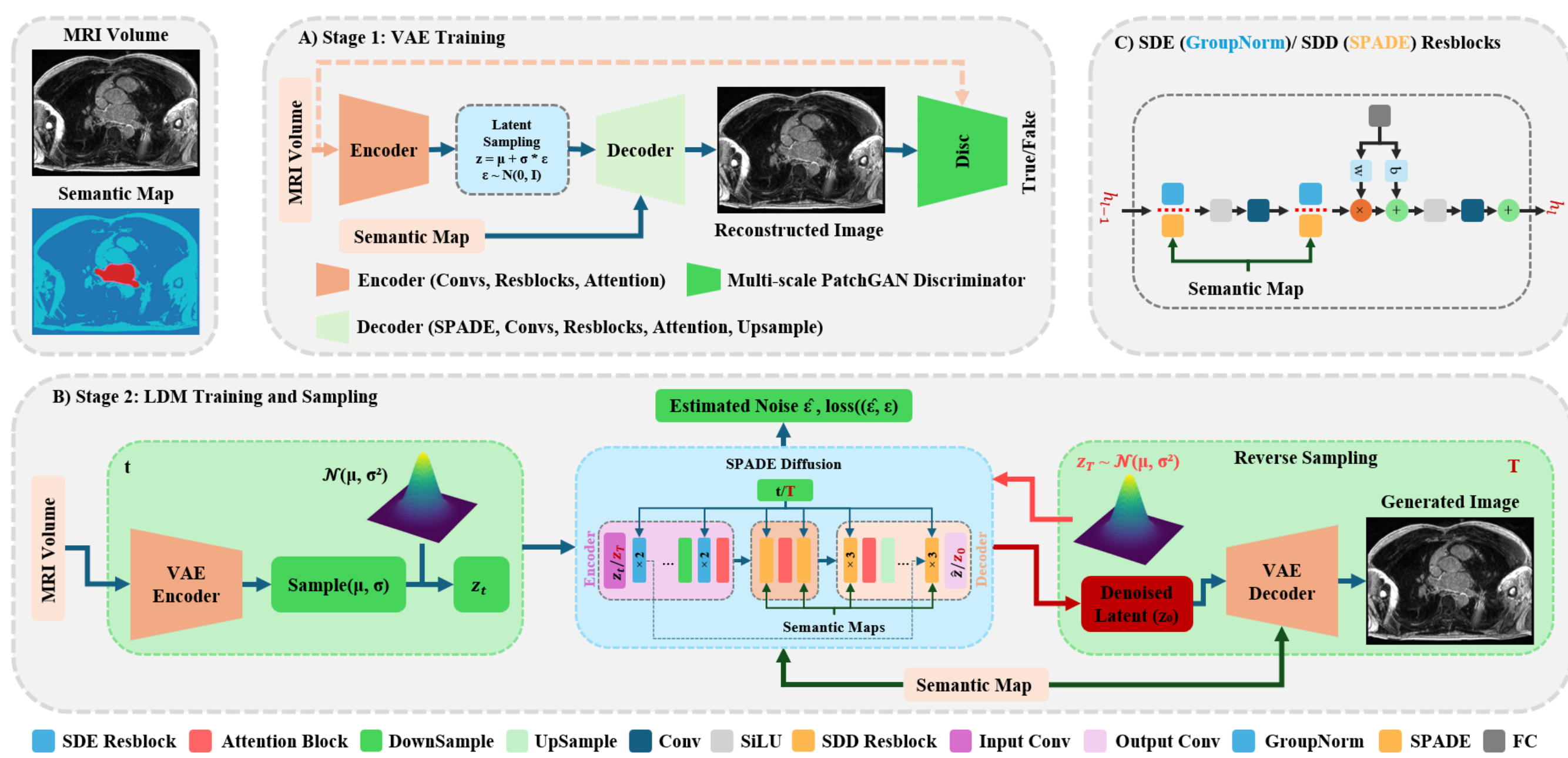

Fig. 1. Overview of SPADE LDM framework. (A) VAE training with SPADE-conditioned decoding for semantic reconstruction. (B) Latent diffusion guided by semantic maps to generate MRI images. (C) SDE/SDD (Semantic Diffusion Encoder/Decoder) ResBlocks in the diffusion model.

### D. Image Quality Metrics

We evaluate synthetic image quality using both distributional and voxel-wise metrics.

Fréchet Inception Distance (FID) and Maximum Mean Discrepancy (MMD) evaluate global distributional similarity in the feature space of a pre-trained 3D ResNet-34 (MedicalNet), while Peak Signal-to-Noise Ratio (PSNR) and Multi-Scale Structural Similarity Index (MS-SSIM) measure voxel-wise fidelity and perceptual quality.

FID computes the Fréchet distance between multivariate Gaussians fitted to real and generated features as in (3):

$$FID = \|\mu_r - \mu_g\|^2 + Tr(\Sigma_r) + Tr(\Sigma_g) - 2\,Tr\left((\Sigma_r \Sigma_g)^{0.5}\right) \tag{3}$$

where $\mu_r$, $\Sigma_r$ and $\mu_g$, $\Sigma_g$ are the mean and covariance of real and generated features, respectively. $Tr(\cdot)$ denotes the matrix trace operator [19].

MMD (4) estimates the distance between feature distributions using a kernel:

$$MMD^2 = \frac{1}{n(n-1)} \sum_{i \neq j} \mathcal{K}(x_i, x_j) + \frac{1}{m(m-1)} \sum_{i \neq j} \mathcal{K}(y_i, y_j) - \frac{2}{nm} \sum_{i,j} \mathcal{K}(x_i, y_j) \tag{4}$$

where $x_i$ and $y_j$ are features from real and generated images, and $\mathcal{K}(\cdot,\cdot)$ is a kernel (dot-product). A smaller $MMD^2$ indicates better alignment between the two distributions [20].

PSNR quantifies reconstruction fidelity as in (5):

$$PSNR = 10 \cdot log_{10}\left(\frac{L^2}{MSE}\right) \tag{5}$$

where $MSE = \frac{1}{N}\sum_{i=1}^{N}\left(I_i - \hat{I}_i\right)^2$ . $I_i$ and $\hat{I}_i$ are real and generated voxel intensities, N is the number of voxels, and L is the dynamic range.

MS-SSIM (6) measures perceptual similarity by combining luminance, contrast, and structural terms across scales:

$$MS-SSIM = \prod_{j=1}^{M-1} CS_j(x,y)^{\beta_j} \cdot SSIM_M(x,y)^{\beta_M} \tag{6}$$

The SSIM at the coarsest scale is computed as:

$$SSIM(x,y) = \frac{(2\mu_x\mu_y + C_1)(2\sigma_{xy} + C_2)}{(\mu_x^2 + \mu_y^2 + C_1)(\sigma_x^2 + \sigma_y^2 + C_2)},$$

The contrast-structure comparison is given by:

$$CS_j(x,y) = \frac{2\sigma_{xy} + C_2}{\sigma_x^2 + \sigma_y^2 + C_2},$$

where $\mu_x, \mu_y$ are the local means and $\sigma_x^2, \sigma_y^2$ are the local variances, $\sigma_{xy}$ is the local cross-covariance. $C_1 = (k_1 \cdot L)^2$, $C_2 = (k_2 \cdot L)^2$ are stabilizing constants, with L being the dynamic range of voxel intensities (1.0), $k_1$=0.01, $k_2$=0.03. $\beta_j$ denotes the scale weight at resolution level $j$ (0.0448, 0.2856, 0.3001), and M =3 is the number of scales [21]. Since these metrics assess fidelity and realism (not generalization), we compute them on the full dataset for robust, representative evaluation, as recommended in prior studies [19].

## III. RESULTS AND DISCUSSION

### A. Composite Semantic Label Maps

We evaluated k-means clustering over $k$=2 to 10 using the Silhouette score (measures cluster separation, higher is better) and Davies–Bouldin Index (DBI), as shown in Fig. 2. The Silhouette score peaked at $k$=3 and remained stable up to $k$ =6. In contrast, DBI (evaluates compactness and separation; lower is better) has minimum values with $k$=2, 3, and 5. Based on this trade-off, we restricted $k$ to the range of 2 to 5 to ensure clustering quality and consistency.

For this study, we used k = 2 to construct composite semantic label maps for conditioning the generative models; exploring higher values of k remains a direction for future work.

Fig. 3 shows sample output from our composite label map generation pipeline. The composite label maps (C, D) are generated from the original MRI (A) and ground truth masks (B). Though approximate, the auxiliary labels introduced contextual diversity and improved synthesis realism, as shown by refinements from k = 2 to k = 5 in (C, D).

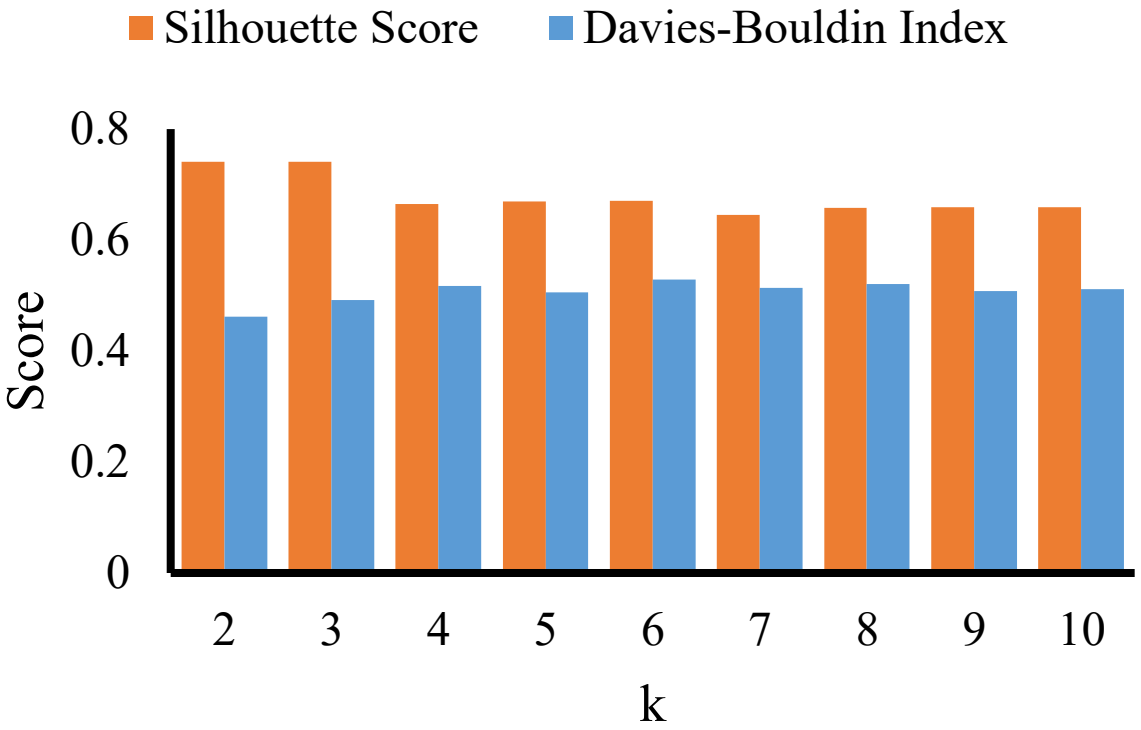

Fig. 2. Silhouette Score and Davies–Bouldin Index for k-means, σ=1.

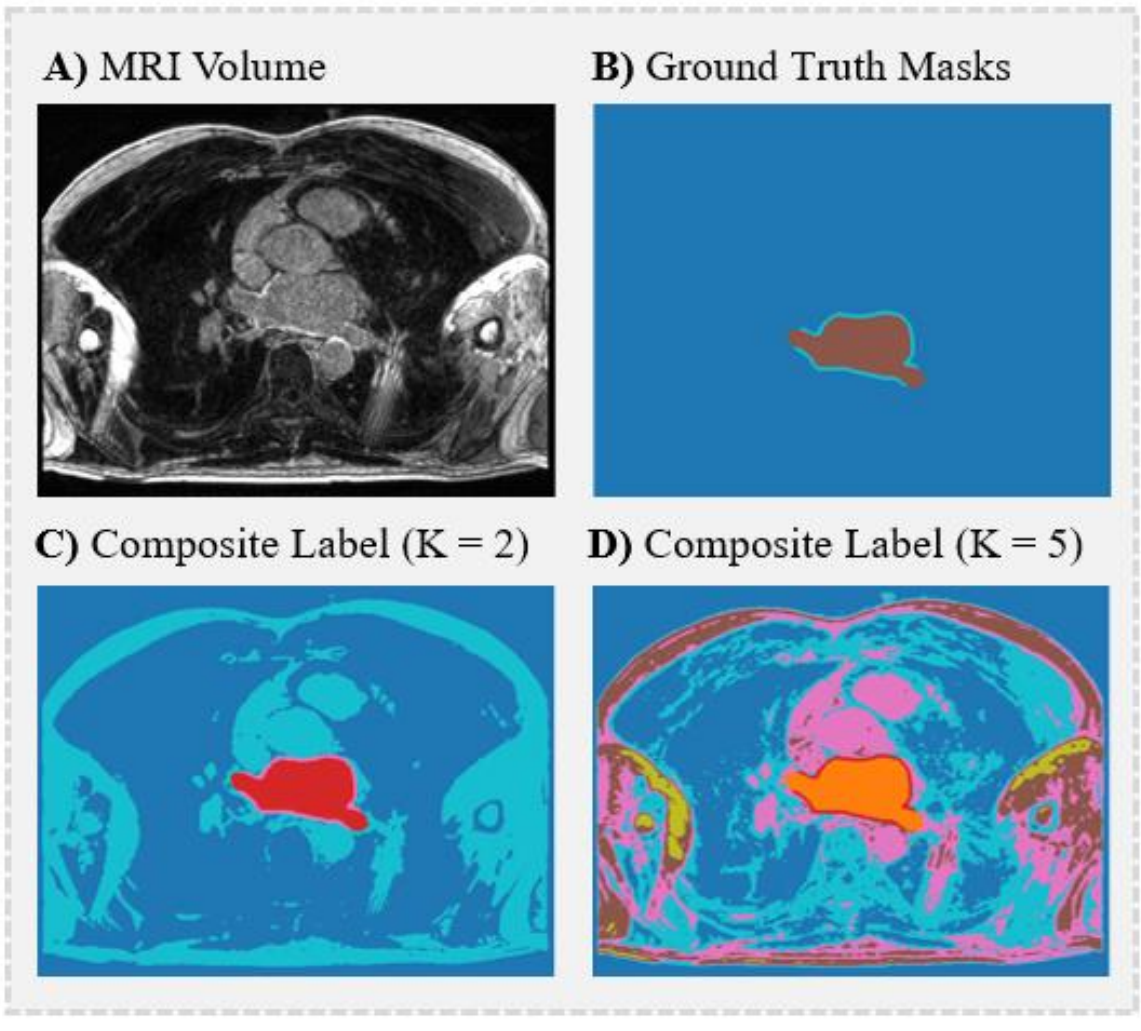

Fig. 3. Sample output from the proposed composite label map pipeline. (A) Original MRI volume; (B) Ground truth masks (endocardium and wall); (C and D) composite label maps with k = 2 and 5 respectively.

## B. *Synthetic Image Quality and Fidelity*

Fig. 4 illustrates the progression of visual quality across models. 3D Pix2Pix captures general LA morphology but produces over-smoothed textures and uniform backgrounds. SPADE-GAN improves the anatomical structure and local texture but occasionally introduces non-physiological speckle artifacts. SPADE-LDM preserves wall detail, contrast dynamics, and fine-grained intensity variation resembling acquisition noise. Although none of the models fully reproduce the anatomical fidelity and visual characteristics of real LGE MRI, the improvement from Pix2Pix to SPADE-GAN to SPADE-LDM demonstrates increasing anatomical fidelity.

When trained only on LA blood pool and wall masks, models reconstructed the labelled regions but failed to synthesize realistic surrounding anatomy. This limitation was mitigated by introducing composite semantic maps, which enabled the generation of more coherent cardiac structures beyond the labelled areas. As shown in Fig. 5, only the output conditioned on composite semantic map (C) closely resembles the real MRI (A), whereas using ground-truth masks alone (B) results in anatomically incomplete synthesis (D).

Table 1 summarizes the quantitative performance of the three models across FID, MMD, MS-SSIM, and PSNR. Lower FID and MMD reflect better realism via feature-space similarity, while higher MS-SSIM and PSNR indicate stronger structural and voxel-wise fidelity. SPADE-LDM consistently outperformed the others, achieving the lowest FID (4.063) and MMD (2.656), indicating close distributional alignment with real LGE scans. It also yielded the highest MS-SSIM (0.826) and PSNR (24.792 dB), confirming excellent anatomical preservation and intensity accuracy.

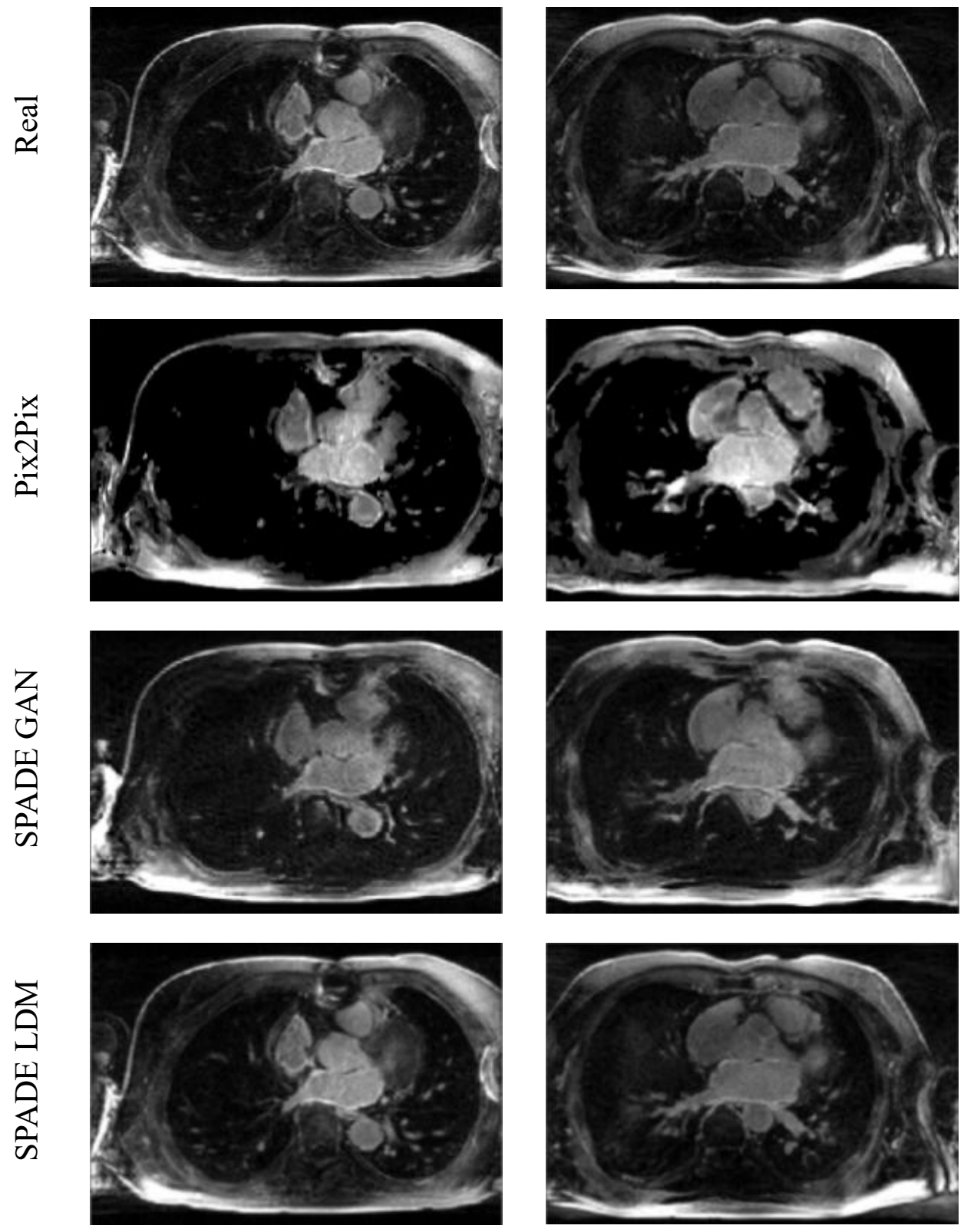

Fig. 4. Comparison of overall quality between real and synthetic 3D LA LGE MRI across axial slices.

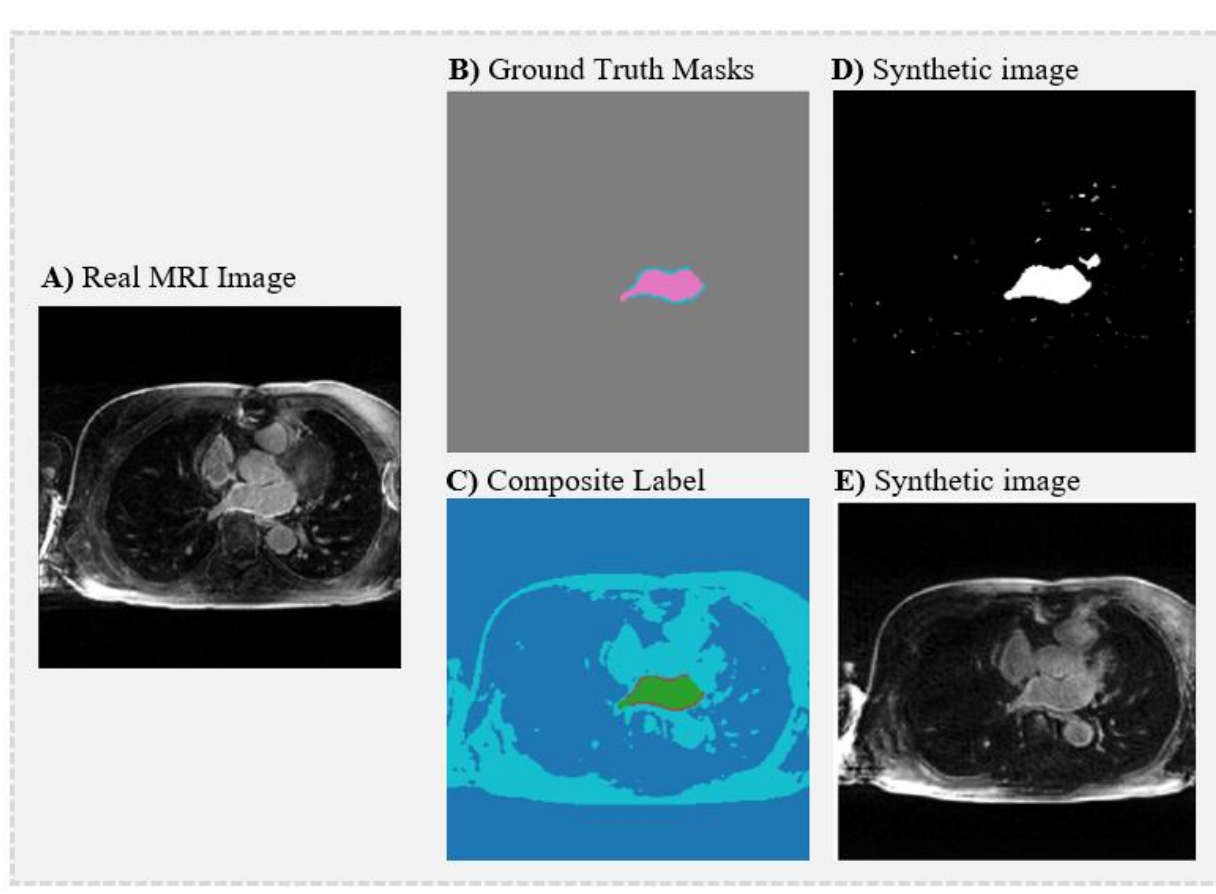

Fig. 5. Sample conditional generation using different label inputs with the SPADE-GAN model. (D) Output conditioned on ground-truth masks (B). (E) Output conditioned on composite semantic map (C), showing closer resemblance to the real MRI (A).

SPADE-GAN also performed well, achieving a relatively low FID (7.652) and MMD (4.433), with MS-SSIM (0.811) and PSNR (23.542 dB), surpassing Pix2Pix on all metrics. However, occasional speckle artifacts slightly compromised its structural fidelity. Pix2Pix, trained with L1 loss and a U-Net generator, showed the weakest realism (FID = 40.821, MMD = 36.890), consistent with its oversmoothed outputs. Still, its MS-SSIM (0.763) and PSNR (23.067 dB) indicate acceptable voxel-wise fidelity. Overall, SPADE-LDM provides the best balance between realism and structural accuracy, underscoring the strength of diffusion models with semantic conditioning in 3D medical image synthesis.

The baseline single-stage 3D U-Net, trained on real data, achieved a Dice score of 0.908±0.0162 on the test set. Augmentation with SPADE-LDM images improved Dice to 0.936±0.0138 with 2.8 percentage points (+3.1% relative). A one-tailed Wilcoxon signed-rank test confirmed statistical significance ($p < 0.05$) for the Dice metric.

These findings support the use of high-quality synthetic images to improve segmentation in LGE LA MRI. Notably, the segmentation model was trained on the LA cavity to evaluate the utility of the generated images; LA cavity segmentation is more tractable than LA wall segmentation and has established benchmarks in prior work. Tailoring a model specifically for LA wall segmentation – an anatomically thinner and more challenging target – remains an important future direction. The cavity segmentation, with a Dice score of ~0.936, also indicates good performance using only a single-stage U-Net, matching the best-published results of 0.932 using two-stage U-Net [1].

Although prior work on generative augmentation for LA LGE MRI is lacking, Lustermans et al. (2022) [5] applied a 2D SPADE GAN to augment LV LGE data and reported a 6% Dice score improvement for scar segmentation. While focused on a different cardiac chamber, their findings support the feasibility of GAN-based augmentation in cardiac MRI.

TABLE 1: SYNTHETIC IMAGE QUALITY COMPARISON

| *Model* | *FID* ↓ | *MMD* ↓ | *MS-SSIM* ↑ | *PSNR (dB)* ↑ |
|---|---|---|---|---|
| Pix2Pix | 40.821 | 36.890 | 0.763 | 23.067 |
| SPADE GAN | 7.652 | 4.433 | 0.811 | 23.542 |
| SPADE-LDM | **4.063** | **2.656** | **0.826** | **24.792** |

Our study extends this line of inquiry to the more anatomically complex left atrium, achieving a 3.1% relative improvement in segmentation accuracy using 3D SPADE-LDM-generated data.

GANs (Pix2Pix, SPADE) enable faster training and inference; however, Pix2Pix is prone to mode collapse (e.g., repetitive textures) and requires more data for training. To address this, we applied extensive offline data augmentation to enhance diversity and mitigate overfitting. In contrast, SPADE-LDM produces diverse samples with varying noise patterns, which likely regularizes the segmenter and enhances robustness. Limitations include using a 3D ResNet pretrained on general medical images for FID computation; cardiac-specific networks or radiologist scoring could yield more anatomically-relevant realism assessments. Our models also lacked shape variation control, as conditioning was limited to the provided masks available. Future work could explore mask morphing or variational mask synthesis to expand anatomical diversity. Additionally, future directions may extend the composite labelling strategy by exploring higher k-values (3–5) for clustering to enrich contextual anatomy and by extending the framework to direct LA wall segmentation.

# IV. CONCLUSION

We presented a comparative study of 3D generative models for synthesizing LGE MRI of the left atrium from semantic label maps. SPADE-LDM produced the most realistic and anatomically coherent volumes when compared to Pix2Pix and SPADE-GAN. Augmenting the training set with SPADE-LDM-generated images led to improvements in segmentation accuracy. The use of composite semantic maps further enhanced anatomical realism and contextual consistency. These findings demonstrate the potential of conditional generative models (particularly diffusion-based architectures) for data augmentation in cardiac MRI applications where labelled data is limited or imbalanced.

# ACKNOWLEDGMENT

This research was enabled in part by the Natural Sciences and Engineering Research Council of Canada (NSERC) Discovery Grant, and by computational resources provided by the Digital Research Alliance of Canada (https://alliancecan.ca).

# REFERENCES

[1] L. Li, V. A. Zimmer, J. A. Schnabel, and X. Zhuang, "Medical image analysis on left atrial LGE MRI for atrial fibrillation studies: A review," *Medical Image Analysis*, vol. 77, p. 102360, Apr. 2022, doi: 10.1016/j.media.2022.102360.

[2] Z. Xiong et al., "A global benchmark of algorithms for segmenting the left atrium from late gadolinium-enhanced cardiac magnetic resonance imaging," *Medical Image Analysis*, vol. 67, p. 101832, Jan. 2021, doi: 10.1016/j.media.2020.101832.

[3] K. Jamart, Z. Xiong, G. D. Maso Talou, M. K. Stiles, and J. Zhao, "Mini review: Deep learning for atrial segmentation from late gadolinium-enhanced MRIs," *Frontiers in Cardiovascular Medicine*, vol. 7, p. 522088, May 2020, doi: 10.3389/fcvm.2020.00086/bibtex.

[4] H. C. Shin *et al.*, "Medical image synthesis for data augmentation and anonymization using generative adversarial networks," in *Simulation and Synthesis in Medical Imaging* (SASHIMI 2018), A. Gooya, O. Goksel, I. Oguz, and N. Burgos, Eds., Lecture Notes in Computer Science(), vol. 11037, Cham: Springer Verlag, 2018, pp. 1–11. doi: 10.1007/978-3-030-00536-8_1.

[5] D. R. P. R. M. Lustermans, S. Amirrajab, M. Veta, M. Breeuwer, and C. M. Scannell, " Optimized automated cardiac MR scar quantification with GAN - based data augmentation, " *Computer Methods and Programs in Biomedicine,* vol. 226, , p. 107116, Nov. 2022, doi: 10.1016/j.cmpb.2022.107116.

[6] M. Mirza and S. Osindero, "Conditional generative adversarial nets," *arXiv preprint* arXiv:1411.1784, Nov. 2014.

[7] P. Isola, J.-Y. Zhu, T. Zhou, and A. A. Efros, "Image-to-image translation with conditional adversarial networks," in *Proceedings of the IEEE Conference on Computer Vision and Pattern Recognition*, 2017, pp. 1125–1134.

[8] T. Park, M. Y. Liu, T. C. Wang, and J. Y. Zhu, "Semantic image synthesis with spatially-adaptive normalization," in *Proceedings of the IEEE Computer Society Conference on Computer Vision and Pattern Recognition (CVPR), Long Beach, CA, USA*, 2019, pp. 2332–2341. doi: 10.1109/cvpr.2019.00244.

[9] J. Ho, A. Jain, and P. Abbeel, "Denoising diffusion probabilistic models," in *Advances In Neural Information Processing Systems*, vol. 33, 2020, pp. 6840–6851.

[10] Z. Dorjsembe, H. K. Pao, S. Odonchimed, and F. Xiao, "Conditional diffusion models for semantic 3D brain MRI synthesis," *IEEE Journal of Biomedical and Health Informatics*, vol. 28, no. 7, pp. 4084–4093, Jul. 2024, doi: 10.1109/jbhi.2024.3385504.

[11] R. Rombach, A. Blattmann, D. Lorenz, P. Esser, and B. Ommer, "High-resolution image synthesis with latent diffusion models," in *2022 IEEE/CVF Conference on Computer Vision and Pattern Recognition (CVPR)*, New Orleans, LA, USA, Dec. 2022, pp. 10674–10685. doi: 10.1109/CVPR52688.2022.01042.

[12] W. Zhou et al., "Semantic image synthesis via diffusion models," *arXiv preprint* arXiv:2207.00050, Jun. 2022.

[13] T. Ruschke et al., "Guided Synthesis Of Labeled Brain MRI data using latent diffusion models for segmentation of enlarged ventricles," *arXiv preprint* arXiv:2411.01351, Nov. 2024.

[14] L. Zhu et al., "Make-a-volume: leveraging latent diffusion models for cross-modality 3D brain MRI synthesis," in *International Conference on Medical Image Computing and Computer-Assisted Intervention-MICCAI 2023*, *Lecture Notes in Computer Science*, vol. 14229, Cham, Switzerland: Springer, Oct. 2023, pp. 592–601. doi: 10.1007/978-3-031-43999-5_56.

[15] S. Amirrajab et al., "XCAT-GAN for synthesizing 3D consistent labeled cardiac MR images on anatomically variable XCAT phantoms," in *Medical Image Computing and Computer Assisted Intervention–MICCAI 2020, Lecture Notes in Computer Science,* vol. 12264, Cham, Switzerland: Springer, Oct. 2020, pp. 128–137. doi: 10.1007/978-3-030-59719-1_13.

[16] S. Amirrajab, Y. Al Khalil, C. Lorenz, J. Weese, J. Pluim, and M. Breeuwer, "Label-informed cardiac magnetic resonance image synthesis through conditional generative adversarial networks," *Computerized Medical Imaging and Graphics*, vol. 101, p. 102123, Oct. 2022, doi: 10.1016/j.compmedimag.2022.102123.

[17] G. Baldini, M. Schmidt, C. Zäske, and L. L. Caldeira, "MRI scan synthesis methods based on clustering and Pix2Pix," in *Artificial Intelligence in Medicine – AIME 2024, Lecture Notes in Computer Science*, vol. 14845, Cham, Switzerland: Springer, Jul. 2024, pp. 109–125, doi: 10.1007/978-3-031-66535-6_13.

[18] Z. Zhang, L. Yang, and Y. Zheng, "Translating and segmenting multimodal medical volumes with cycle-and shape-consistency generative adversarial network," in *2018 IEEE/CVF Conference on Computer Vision and Pattern Recognition (CVPR)*, Salt Lake City, UT, USA, 2018, pp. 9242–9251. doi: 10.1109/cvpr.2018.00963.

[19] M. Heusel, H. Ramsauer, T. Unterthiner, B. Nessler, and S. Hochreiter, "GANs trained by a two time-scale update rule converge to a local Nash equilibrium," in *Advances in Neural Information Processing Systems*, vol. 30, Long Beach, CA, USA, 2017, pp. 6629–6640. doi: 10.18034/ajase.v8i1.9.

[20] A. Gretton, K. M. Borgwardt, M. J. Rasch, B. Schölkopf, and A. Smola, "A kernel two-sample test," *The Journal of Machine Learning Research*, vol. 13, pp. 723–773, 2012.

[21] Z. Wang, E. P. Simoncelli, and A. C. Bovik, "Multi-scale structural similarity for image quality assessment," in *The Thirty-Seventh Asilomar Conference on Signals, Systems & Computers*, 2003, Pacific Grove, CA, USA, 2003, pp. 1398–1402. doi: 10.1109/acssc.2003.1292216